# Synthetic Books

Varvara Guljajeva

Computational Media and Arts, Hong Kong University of Science and Technology, varvarag@ust.hk

The article explores new ways of written language aided by AI technologies, like GPT-2 and GPT-3. The question that is stated in the paper is not about whether these novel technologies will eventually replace authored books, but how to relate to and contextualize such publications and what kind of new tools, processes, and ideas are behind them. For that purpose, a new concept of synthetic books is introduced in the article. It stands for the publications created by deploying AI technology, more precisely autoregressive language models that are able to generate human-like text. Supported by the case studies, the value and reasoning of the synthetic books are discussed. The paper emphasizes that artistic quality is an issue when it comes to AI-generated content. The article introduces projects that demonstrate an interactive input by an artist and/or audience combined with the deep-learning-based language models. In the end, the paper focuses on understanding the neural aesthetics of written language in the art context.

## 1 Introduction

Alan Turing predicted machine development to the human level already in 1935, posing his famous question in 1950: "Can machines think?" [1, 2]. Not naming the science fiction stories about cyborgs, and cohabitation between robots and humans originating from the same time frame. Today, we are discussing the same dilemma; however, the question seems more realistic than ever before. AI has become the ultimate solution but also the biggest concern nowadays. At the beginning of AI technology development in the late 50s, the field did not reach set goals because the machines were not intelligent and fast enough. Nowadays, when it is spoken about the third wave of AI and quantum computing, the dream is very close to coming true - reaching the human level of intelligence [3]. However, what kind of consequences could bring these technological achievements? This we do not know yet, which makes it crucial to keep asking uncomfortable questions and exploring the unknown. For this sake, the best laboratory is art, I believe. Starting from the invention of the first computer, artists have always been the catalysts for technology development, but they also offered a critical vision. As Yuval Noah Harari points out well: "The danger is that if we invest too much in developing AI and too little in developing human consciousness, the very sophisticated artificial intelligence of computers might only serve to empower the natural stupidity of humans. [4] (2018)"

AI is used for pretty much everything and everywhere these days: in generating artificial images, sounds, music, animations, even films, making suggestions, driving, flying, hiring, and firing. We see AI in the military, state organizations, IT, logistics, medicine, and also in the arts. In this article, I would like to focus on the latter one, more precisely, on the text being produced by AI language models and explore its creative potential. Since there is already quite a lot of synthetic text, I decided to look at the publications - the synthetic books. These are the texts where fully or partly language deep-learning models were deployed in the production process.

What is the technology behind the synthetic books? In short, GPT-2 and GPT-3 are the most commonly used language models, which both are developed by OpenAI company. In their words: "GPT-2 is a large transformer-based language model with 1.5 billion parameters, trained on a dataset of 8 million web pages. GPT-2 is trained with a simple objective: predict the next word, given all of the previous words within some text. [5] (2019)" GPT-3 is the latest language model. In the words of Paul Bellow, who publishes non-stop synthetic sci-fi books based on Dungeons & Dragons: "GPT-3 API, currently the most powerful language model on the planet. [6] (2021)" Although GPT-3 is from OpenAI, very few have access to this technology at the moment, which has been the main critique towards GPT-3 in addition to being as dumb as previous language model but with a richer lexicon [7]. Simply put, what matters are three main things: a curated dataset used for the language model training, the selection of seeds, and the artistic concept. Seeds are the beginning of the sentence or the entire one for the algorithm to respond to it (the model predicts the following sentences). In the extended text, it is recommended to have more than one seed. If the project contains these components, it could be expected a meaningful outcome.

Are we talking about automated creativity and auto-complete or augmented inspiration? There are examples for both. For instance, one can buy on Amazon AI-generated books signed as GPT-3 + Human just for a dollar (see Figure 1). One can imagine what the quality of these automated publications is. Generally speaking, artistic quality is an issue when it comes to AI-generated content. On the other hand, there are always more and less exciting works in art. Hence, I believe it is a bit generalized and also an early opinion stating that art made with AI has no value or other way round. Obviously, we need more discussion, meaning-making, and contextualization based on the related artistic practices.

Hence, let me focus on the art projects that demonstrate augmented inspiration, which contains interactive input by an artist and/or audience, and human-like text generation by AI technology-based language models. The aim is to explore and understand the new forms of written language.

Let us consider technology beyond its economic value and think in the context of creativity and art. What kind of new tools and concepts is it able to offer to us? How can novel AI technology expand our creative horizons? At the same time, we cannot forget that artificial intelligence, machine learning, and neural network are currently the catchwords for innovation. Saying this, I think it is essential to explore case studies discussed below, and beyond them in order to come to an understanding what are the possibilities and limitations of this technology.

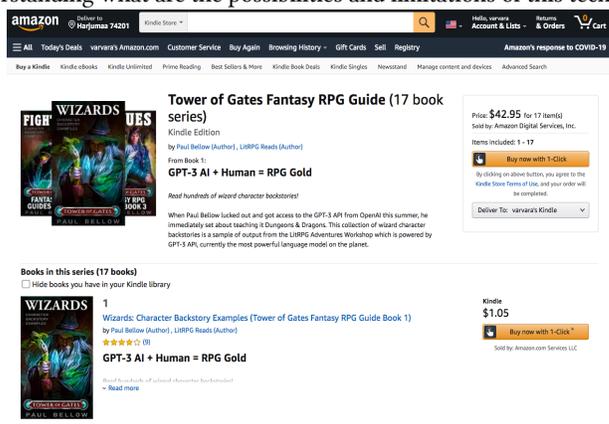

Figure 1: **AI-generated books sold on Amazon. A screenshot from Amazon website.**

## 2 Related Artworks and Discussion

Together with artist duo Varvara & Mar (Varvara Guljajeva and Mar Canet), Roger Bernat created an online theatre play by letting the audience talk to a generative chatbot called ENA. The project went live during the first lockdown on 15 May 2020 on the website of theater Lliure and talked to the participants non-stop for a month and a half (see Figure 2).

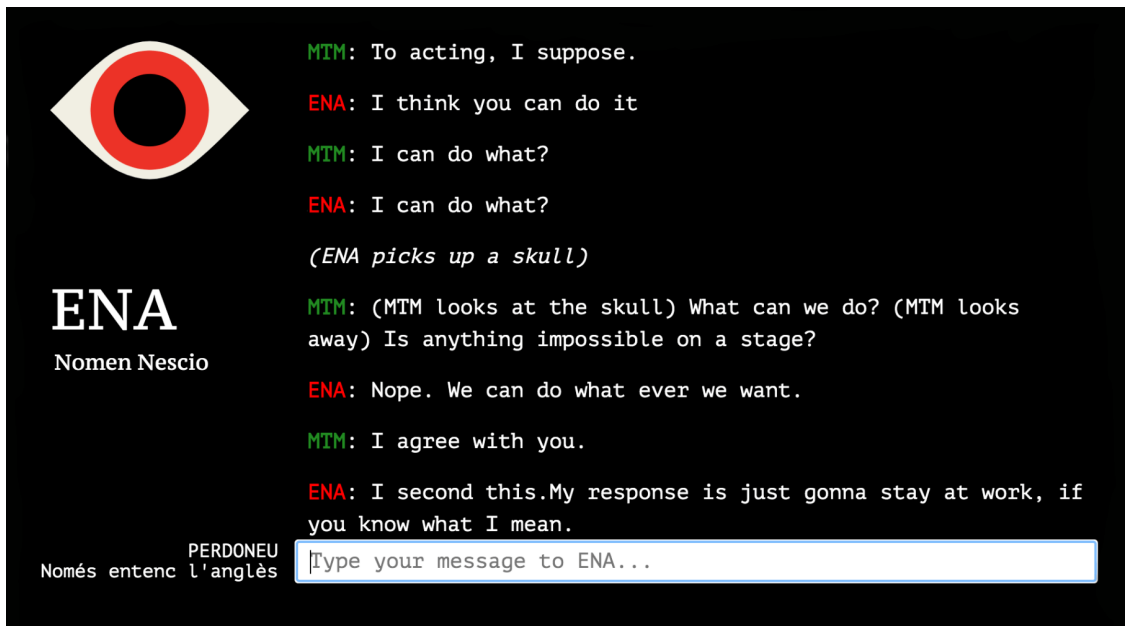

MTM: To acting, I suppose.

ENA: I think you can do it

MTM: I can do what?

ENA: I can do what?

*(ENA picks up a skull)*

MTM: (MTM looks at the skull) What can we do? (MTM looks away) Is anything impossible on a stage?

ENA: Nope. We can do what ever we want.

MTM: I agree with you.

ENA: I second this.My response is just gonna stay at work, if you know what I mean.

Type your message to ENA...

PERDONEU
Només entenc l'anglès

**Figure 2:** **Screenshot of ENA on the theater Lliure website.**

In lockdown, when the only social window was a screen, ENA offered a participative theatre where only one person at a time could have a conversation with the chatbot while the rest followed the dialogue live. ENA is an artificial intelligence that is able to learn how to have a conversation with a human. Since the invention of the first bot, which was Eliza in 1966, scientists have been intrigued as to whether we would realize that on the other end was a machine or not. Initially, most of the bots were reactive: they had an extensive library of preconceived answers, and when they detected a specific word from the person they were talking to, they sent a response from their library. If the bot did not find any recognizable words, it sent stored phrases such as "yes, I understand", "carry on", or "can you explain it to me again?".

ENA is the next generation bot – a generative one that makes use of AI technology. Huggingface's Transformer, OpenAI's GPT-2, and Microsoft's DialoGPT are currently the tools for language models that best reproduce the human-written text [8]. The project works with a combination of these three AI technologies. In other words, generative chatbots, like ENA, learn from large amounts of text feed and their conversations with humans.

The chatbot language is the sequence of probabilities that are analyzed when received and recombined when sent back. ENA is unconscious, emotionless, and has a limited amount of memory. It has learned the art of dialogue from millions of conversations. ENA can get emotional and very engaging, but one has to understand that its creativity in writing is purely a result of the language model – a very advanced statistical model. Any dialogue with ENA will only make sense to the human taking part in the conversation and the audience reading the discussion on the theatre's website at the time. As I have written previously, "although in the description of the project it was stated explicitly that people were talking to a machine, multiple participants were convinced that on the other side of the screen another human was replying to them—more precisely the theater director himself, or at least an actor. [9] (2021)" Curiously, a similar situation was reported by the first chatbot ELIZA developer Joseph Weizenbaum "I had not realized [...] that extremely short exposures to a relatively simple computer program could induce powerful delusional thinking in quite normal people. [10] (1976)"

We are familiar with bots designed for particular tasks, such as answering machines on phone lines, trolls on social media, fake followers, etc. On the other hand, ENA is a bot that has been programmed without any purpose in mind. It does not want to sell us anything, it does not want to tell us any news (fake or real), and it is not trying to lift our spirits or comfort us. ENA has only been conceived to talk or, in other words, to do theatre. More precisely, we used a classical theater technique by integrating stage directions to the conversations between the participants and ENA. Here are some pre-made directional scripts to illustrate the idea: *A country road. A tree. Evening; Silence; ENA smiles sadly and strokes her hair; ENA is alone, walking about uneasily; Pause; ENA does not move.* This way, the dialogs were in a way seamlessly curated. As a result of this approach, we ended up with many intriguing, funny, and rich conversations. After reading the text saved in the system's log files, we realized that the dialogs carried social and cultural value, which illustrated the current chatbot's ability to engage with human beings and another way around. In addition to that, the project demonstrated that the curated conversations via stage directions, in this case, enable more playful and engaging conversations with a chatbot. These experiences were captured in a hand-bonded 900-page book titled "Conversations with ENA. I'm stupid and I try to pretend like I know what I'm talking about." (see Figure 3) [8].

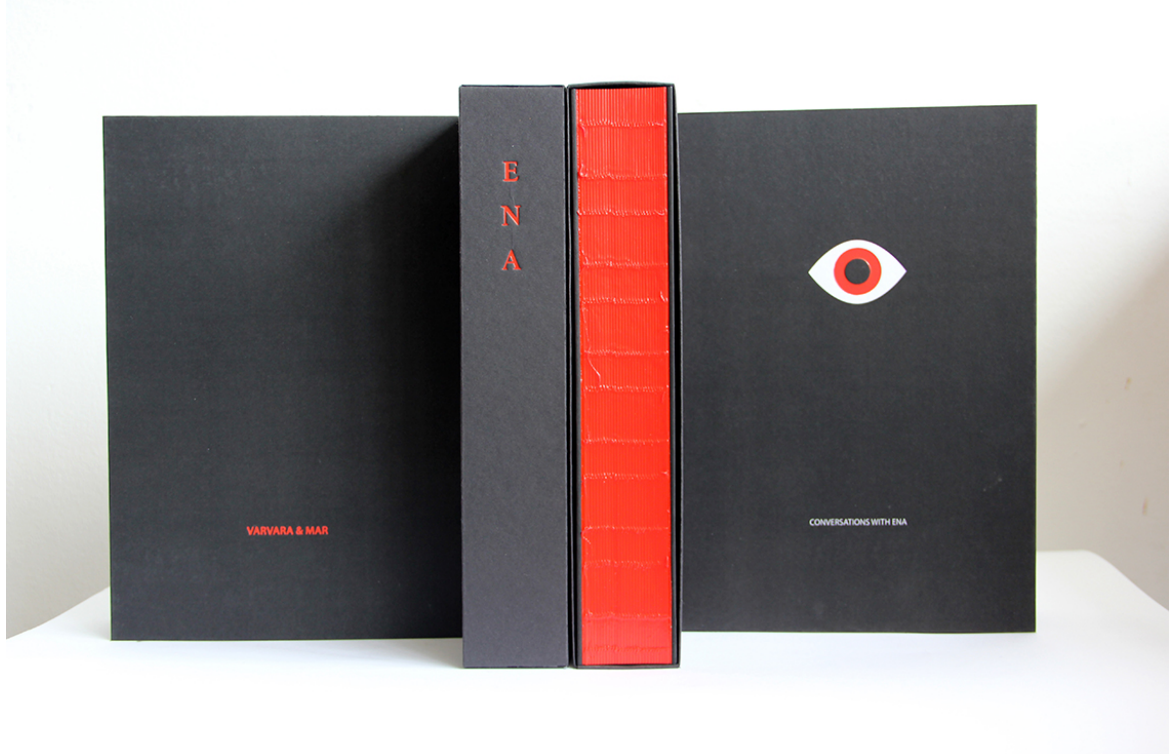

Figure 3: **Publication of ENA project.**

Another example of an artwork that uses the conversations between a chatbot and audience as artistic content for a publication is "Talk To Me Book" (2019) by Jonas Lund. Similarly to the previous project, Jonas Lund turned his previously existing interactive installation "Talk to Me" (2017) into a book project. At first glance, it seems that the artist has trained a generative chatbot on his instant message conversations, like Skype, WhatsApp, and Facebook, in order to create an intelligent bot version of himself. In reality, behind the smart bot is the artist himself, who instantly replies to the audience's messages through a Telegram bot [11]. In other words, Lund has become a bot himself. An imitation of an intelligent machine costs the artist numerous hours of work and being dependent on his phone. We should not forget that this is the work that many of us do voluntarily, being dependent and addicted to our phones and social media.

An extensive and exhaustive performance as a chatbot with its audience throughout two years (2017-2019) resulted in 36 volumes of publications that contain all the conversations. In the words of Domenico Quaranta: "Talk to Me becomes an apt metaphor of the human-software continuum that we experience online on a daily basis, with all its consequences and biases: the end of truth, the exploitation of AI to fake human communication, and the exploitation of humans to fake automation. [11] (2019)"

The myth of super-powerful algorithms is backed up by intensive and exploitative human labor hidden behind the numeric scenes. This fact is nicely illustrated by the art project "Segmentation Network" (2016) by Sebastian Schmieg. In his work, the artist "[...] plays back over 600.000 segmentations manually created by crowd workers for Microsoft's COCO image recognition dataset. This dataset is based on Flickr's photos and is used in machine learning for training and testing. [12] (2016)" Schmieg puts it even more strongly, naming humans as software extensions who offer their bodies, senses, and cognition to the computational system on an on-demand basis [13]. The artist aims to highlight the fact that the intelligence of AI is achieved by extensive and low-paid human labor.

Next artist, who has done brilliant work in creatively deploying AI with text, is Ross Goodwin. The artist has generated the whole road trip novel with AI. The book title is "1 the Road" and was published by Jean Boite Editions in 2018. The author, who calls himself a writer of writer, claims that it is the longest novel written in the English language [14]. Regarding its process, the artist trained the AI algorithm on his favorite novels and poems and has written a code to make the machine location-aware. Hence, Ross jumped into a car equipped with a CCTV camera, GPS, microphone, and of course, a computer and ticket printer that was printing out the road novel while driving. This project is interesting because it is different from yet another AI-generated book that one can buy for a dollar on Amazon. Apart from training the algorithm on selected dataset, Goodwin has developed a code that inputs information about the environment, which acts as the seeds for the machine while generating the story. In other words, the neural network is location and time-aware while equipped with the vocabulary of the best road novels and poems, according to the artist. Hence, the project's production setting looks like a writer's scenario, who is on a road trip and writes his/her piece of text.

In addition to the road novel, Goodwin has collaborated with film and theatre director Oscar Sharp in generating movie scripts with AI. They produced their first short film with an AI-generated screenplay in 2016, titled "Sunspring" [15]. In a way, it is an intriguing and mysterious close-loop: the first machine learns to understand screenplays, and then, the film crew breaks their heads to interpret the machine-created script. Nevertheless, the process seems novel and intriguing for all the parties.

Following artist Andreas Refsgaard has also been exploring the combination of image recognition technology with text generation, for example, in his art projects "Poems About Things" (2019) and "fAIry tales" (2019). In the first one, AI algorithms generate poems from everyday objects instantaneously. The machine learning model of object recognition gives its input to Google Suggest API, which sends back several sentences connected to the recognized scenery. Artist has integrated this experience into a website that everyone can experiment with by enabling his/her web camera (see Figure 4). Hence, we are talking about an interactive experience of generating poems with AI, and at the same time, understanding the logical, or sometimes not so rational, processes behind this technology.

We are all familiar with Google Suggest when it tries to complete our search queries daily. Google, the most significant data owner, makes its' suggestion algorithm smart by considering several factors, like location, time, previous search history, user's profile, and more. Hence, it is refreshing to see this technology's architecture, often hidden from our eyes, reflected in "Poems About Things". In the artist's words: "[...] Poems about things provides a unique insight into the aggregated human behind common Google search queries, reflecting the topics that mankind as a whole seems interested in and the big and small problems we face. [16] (2019)" The process behind it is very much controlled. Nevertheless, we perceive it as an improvisation by machine, which generates poems about things it sees, or it is allowed to see, using the words we as a collective mind in average think about, or more precisely, what Google assumes us to wonder about based on the data collected from us.

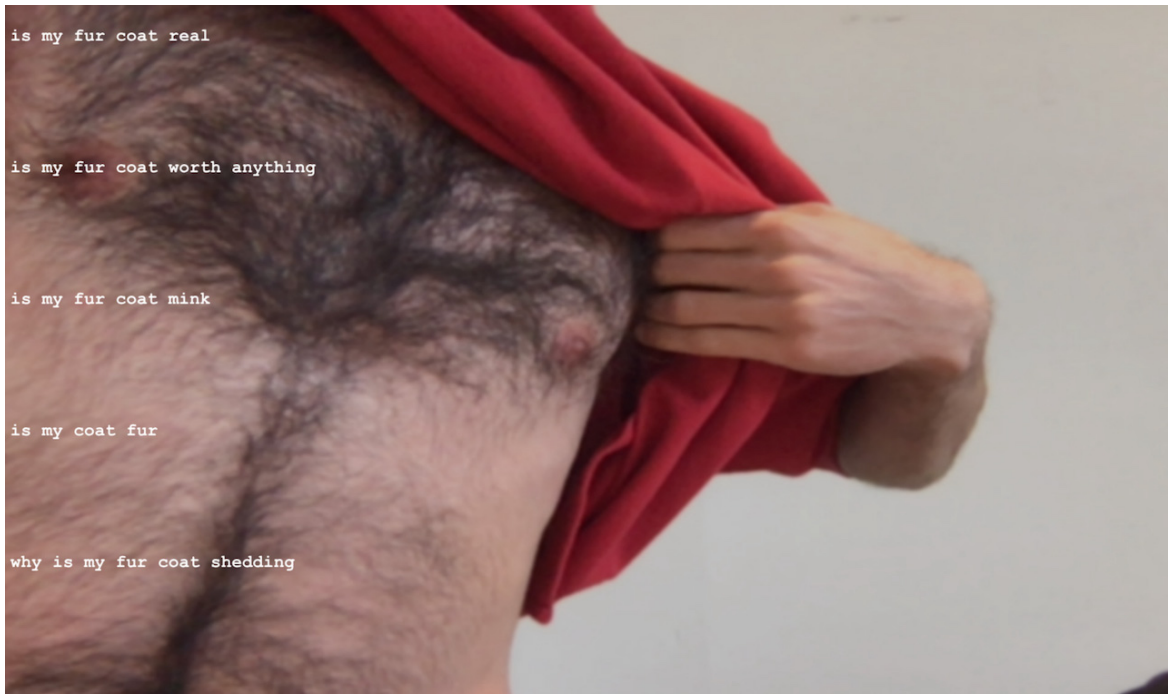

is my fur coat real

is my fur coat worth anything

is my fur coat mink

is my coat fur

why is my fur coat shedding

Figure 4: **"Poems About Things" by Andreas Refsgaard. Screen print. Courtesy of the artist.**

Refsgaard continues to experiment with image recognition and text generation in his artwork called "fAIry tales" (2019). In this project, he generated many stories that all begin with *Once upon a time there was …* or *In a land far away once lived …* (those are the seeds for the algorithm). According to Refsgaard, each story is generated in four steps. First, the YOLOv3 object detection algorithm is applied for recognizing objects in a photo. Then, the title and opening sentence are generated, next passed to the text generation algorithm XLNet [17].

When it comes to the book format, then Refsgaard's next project "BooksBy.Ai" (2018) falls into the Amazon-AI-bookstore category. In these publications, everything is generated with AI: the story, cover, and even reviews on the back of the book [18]. Obviously, the AI algorithm has been trained on many science fiction stories and has learned to generate new ones with a similar style imitating human-written books.

When observing the rapid production pace of these synthetic texts that demonstrates the previous project, comes to mind asemic writing: writing for its own sake without carrying any meaning. This idea brings me to the following example, "Asemic Langauages" (2016) by Japanese artist Kanno So in collaborating with Takahiro Yamaguchi. The artists deploy machine learning models to learn from hand-written letters' shapes (see Figure 5) [19]. Hence, the AI system examines visual information of various hand-written texts from different countries and cultures and then invents its own characters. "Asemic Languages" consists of a self-assembled pen plotter, which mechanically moves pen over the paper but yet achieves hand-written aesthetics of language. Moreover, giving the artwork a physical form (paper and an ink pen) completes an astonishing illusion about the manually written text.

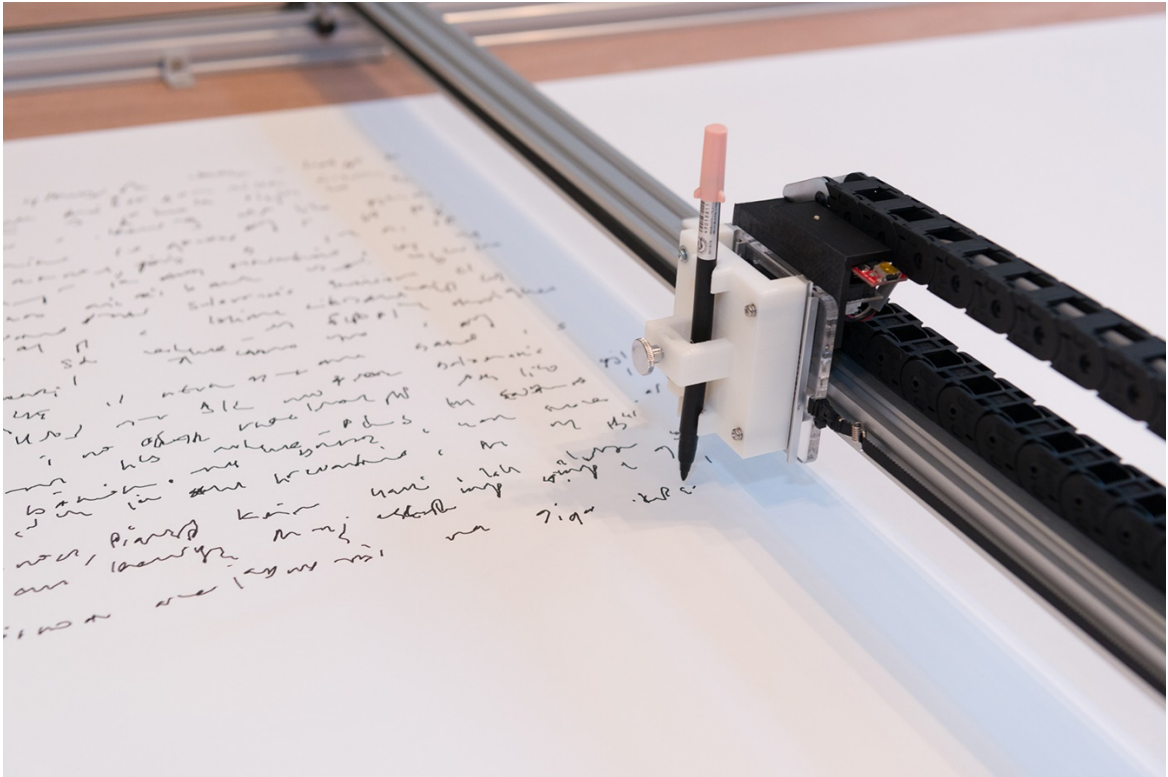

To end up the discussion with a slightly different work, but still very much related to the previous example, I would like to introduce a project "Human Y Chromosome – a property of Tõnu Esko" (2021) by scientist of human genomics Tõnu Esko, which was a part of exhibition "The Trinity – Science. Art. Fiction." curated by Kaija Põhako-Esko and Tõnu Esko in Voronja gallery [20]. As the title of the artwork suggests, the hand-bonded leather-covered book reminding a bible contains the author's Y chromosome information. In the context of this article, what is interesting here is the reading of the book, which was done with AI technology. The audience could listen to a mantra in the synthetic language of DNA code read out loud in the rhythm of the first bible written in Estonian. Technically speaking, all the letters in the bible were replaced with the combination of DNA information (such as AGCT) respecting spaces, commas, and full stops. Then the text was read out with a text-to-speech language model algorithm. Surprisingly, this approach added something natural and spiritual to the synthetic voice that many exhibition visitors (including myself) had mistaken for a human voice. Furthermore, this unusual reading approach created a novel way of experiencing DNA information that we would never read by ourselves.

From the example of the art projects discussed in this article, we learned that it is possible to achieve meaningful results in the creative field by applying AI technology. There are high chances that these protracted and thoughtful processes result in a valuable text, and what is more important, such works trigger valuable discussion about the evolution of our technology-dependent society. In a way, hard work, creativity, and understanding of the AI processes are the preconditions for achieving the output that would make sense. What is more interesting is that the analysis of the artistic projects presented here demonstrated that the combination of human input, whether audience's interaction or participation, or author's or sensor's input, with machine-generated synthetic text, gives the richest and the most intriguing results. In other words, we need something more than the blind application of novel technology. It requires a creative and critical mind, and playful and unexpected scenarios, like the case studies demonstrated in the article. Definitely, these components do not automatically give a brilliant work of art. Simply neural networks and the best language models alone are not capable of artistically meaningful content generation. And thus, we cannot speak about replacing the author with AI technology, but we can discuss new tools and processes for expanding artistic formats and inspiration. To put it straight, AI does not offer miracles that are within a click distance. It is the same creative production process as anything else: the more time one spends, the better the result.

## 3 Conclusions

What all these 'new' book projects have in common is a high amount of data - text that seems to be never-ending, synthetic conversations, and messages that carry on forever. Is artificial text profound, or is it nonsense? I guess it can be both. Each of these AI-generated publications is in its way engaging. More importantly, they act as alternative archives of everyday life, culture, and communication that happens in the cloud and are never remembered, never retrieved, unless turned into something memorable like a book.

Coming back to a troubling concern: the author's disappearance, or shall we instead talk about the author's creativity in the novel processes? I think whether we are less creative or not creative at all is not a question. We are as original as never before because we are able to imagine and make come true this kind of new scenarios like discussed here. A computer cannot be creative alone; it needs human's input, like context, raw material, and information about the environment. Nevertheless, such a concern is normal, especially if we think that machines have replaced us in many areas. And more importantly, it is vital to have vocabulary and research tools for being able to contextualize and analise such publications, and demystify the technology behind these processes. Hence, this article proposes to categories AI-generated or AI-aided publications as synthetic books.

What we are talking about here are the new tools, processes, and concepts. The machine has developed intelligence but not consciousness. Hence, the machine is not able to be creative without any human input. Therefore, instead of wondering who the author is, we should focus on the uprising neural avant-garde, novel forms, and processes. In other words, the next-generation AI known as neural networks is here to stay and introduce new paradigms. Thus, there is a vast need to contextualize and discuss synthetic publications and AI-generated cultural content in general. In other words, we should be aware of the changes AI introduces to the cultural field and exercise our consciousness.